\documentclass[runningheads]{llncs}
\usepackage[T1]{fontenc}
\usepackage{graphicx,verbatim}
\usepackage{amsmath}
\usepackage{multirow}
\usepackage{amssymb}
\usepackage{booktabs}
\usepackage[colorlinks=true, urlcolor=blue, linkcolor=blue, citecolor=blue]{hyperref}
\usepackage[table]{xcolor}
\usepackage{marvosym}

\def\eg{\emph{e.g.}}

\def\etal{{\em et al.~}}

\def\ourmodel{HSS-Net}

\begin{document}

\newsavebox\CBox
\def\textBF#1{\sbox\CBox{#1}\resizebox{\wd\CBox}{\ht\CBox}{\textbf{#1}}}

\title{Hierarchical Spatio-temporal Segmentation Network for Ejection Fraction Estimation in Echocardiography Videos}

\author{
    Dongfang Wang \and  
    Jian Yang \textsuperscript{(\Letter)} \and 
    Yizhe Zhang \and 
    Tao Zhou \textsuperscript{(\Letter)}
} 
%index{Wang, Dongfang}
%index{Yang, Jian}
%index{Zhang, Yizhe}
%index{Zhou, Tao}

\authorrunning{D. Wang et al.}

\institute{
    PCA Lab, Key Lab of Intelligent Perception and Systems for High-Dimensional Information of Ministry of Education, School of Computer Science and Engineering, Nanjing University of Science and Technology, Nanjing, China\\
    \email{\{dongfangwang,csjyang\}@njust.edu.cn}
    \email{\{yizhe.zhang.cs,taozhou.ai\}@gmail.com}
}

\titlerunning{\ourmodel~for Echocardiography Video Segmentation}    
\maketitle              

\begin{abstract}

Automated segmentation of the left ventricular endocardium in echocardiography videos is a key research area in cardiology. It aims to provide accurate assessment of cardiac structure and function through Ejection Fraction (EF) estimation. 
Although existing studies have achieved good segmentation performance, their results do not perform well in EF estimation. 
In this paper, we propose a Hierarchical Spatio-temporal Segmentation Network (\ourmodel) for echocardiography video, aiming to improve EF estimation accuracy by synergizing local detail modeling with global dynamic perception. The network employs a hierarchical design, with low-level stages using convolutional networks to process single-frame images and preserve details, while high-level stages utilize the Mamba architecture to capture spatio-temporal relationships. The hierarchical design balances single-frame and multi-frame processing, avoiding issues such as local error accumulation when relying solely on single frames or neglecting details when using only multi-frame data.  
To overcome local spatio-temporal limitations, we propose the Spatio-temporal Cross Scan (STCS) module, which integrates long-range context through skip scanning across frames and positions. This approach helps mitigate EF calculation biases caused by ultrasound image noise and other factors.
We achieved state-of-the-art results on three datasets. 
Our code is available at \url{https://github.com/DF-W/HSS-Net}.


\end{abstract}

\begin{figure*}[t]
\centering
\includegraphics[width=1.0\textwidth]{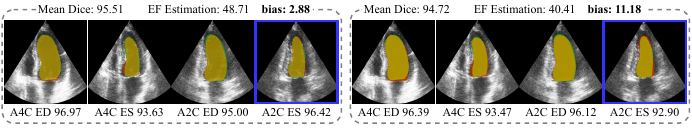} % Reduce the figure size so that it is slightly narrower than the column.
\caption{Left ventricular segmentation maps for ED and ES frames from A2C and A4C views, with Dice segmentation metrics displayed at the bottom. \textbf{It is evident that smaller segmentation errors in the key regions (base and apex) within the blue box lead to greater EF calculation deviations}. The green, red, and yellow represent the ground truth, prediction, and overlapping regions, respectively.}
\label{fig1}
\end{figure*}

\section{Introduction}
The primary of echocardiography analysis is to assess cardiac function, where accurate segmentation of the left ventricular endocardium is essential to measure the heart's Ejection Fraction (EF)~\cite{chen2020deep,akkus2021artificial}.  
However, achieving automatic echocardiography segmentation presents significant challenges. Firstly, the quality of ultrasound images is often compromised by noise, artifacts, and boundary blurring~\cite{10787068,10980661}.
Secondly, the heart undergoes complex motion and deformation during each heartbeat cycle. 

Moreover, physicians typically only annotate the End-Diastolic (ED) and End-Systolic (ES) key frames, resulting in limited and sparse annotation data. 
Therefore, the automated segmentation method is trained on limited annotated data, providing a reliable foundation for the accurate evaluation of cardiac function clinical metrics.

Recently, deep learning methods for echocardiography video segmentation have emerged rapidly. Although effectiveness achieved, their results are not satisfactory when applied to ejection fraction estimation~\cite{thomas_lightweight_2022,SSCF,wu2023super,yang_graphecho_2023}. 
Many studies focus on 2D segmentation of ED and ES frames, with numerous unlabeled frames left unused, resulting in an inability to capture the continuity of cardiac motion~\cite{moradi2019mfp,leclerc2020lu,zamzmi2022real}.  
To address this, Painchaud \etal~\cite{painchaud2022echocardiography} proposed a post-processing framework that leverages cardiac anatomical priors to enhance inter-frame consistency. However, its performance is highly dependent on the quality of the initial segmentation. Similarly, Wei \etal~\cite{wei2023co} used generated pseudo-labels for collaborative learning of segmentation and tracking, while the method is constrained by the quality of the pseudo-labels. 
The Transformer architecture, with multi-head self-attention~\cite{vaswani2017attention}, has been widely adopted in video object segmentation~\cite{arnab2021vivit}. Temporal continuity across frames offers valuable segmentation cues. To exploit this, some methods add specialized modules atop self-attention to capture temporal information. However, this may cause over-reliance on multi-frame relations while overlooking fine details. Moreover, modeling long sequences with many labeled frames introduces heavy computational costs.

In this paper, we propose a Hierarchical Spatio-temporal Segmentation Network (\ourmodel) for echocardiography video segmentation, enhancing the precision of ejection fraction estimation. The low-level stages utilize convolutional networks to process single-frame images, preserving fine details, while the high-level stages employ the Mamba architecture to capture spatio-temporal relationships across multiple frames. By better integrating local and global features, the model reduces segmentation errors in key regions (\eg, the base and apex) and mitigates volume calculation biases caused by local errors, as shown in Fig.~\ref{fig1}.
The network captures inter-frame motion patterns, enhancing the temporal consistency of the segmentation results and improving EF estimation stability. To fully utilize unlabeled data and strengthen the capture of dynamic cues, we propose a spatio-temporal cross scan module, which captures dynamic cues from different spatio-temporal perspectives. This module integrates long-range dependencies, such as apex motion changes and the correlation of lateral wall contraction, through a skip-connections-based spatio-temporal scan. Additionally, global dynamic modeling reduces sensitivity to interference factors, preventing abnormal jitter in segmentation boundaries and ensuring the reliability of EF estimation.

\textbf{Contributions:} \textbf{{\romannumeral 1})} We propose a hierarchical spatio-temporal segmentation framework, which combines convolutional and Mamba architectures for detailed local and temporal processing. %This design alleviates computational burdens in video processing, enabling the model to achieve an optimal balance between single-frame and multi-frame information processing. 
\textbf{{\romannumeral 2})} A STCS Module is proposed to capture dynamic cues from multiple perspectives, enhancing robustness and accuracy. \textbf{{\romannumeral 3})} We present a novel jump scanning mechanism, which breaks local correlations to integrate global information, improving generalization across diverse samples. \textbf{{\romannumeral 4})} Experimental results show the effectiveness of the proposed model.

\begin{figure*}[t]
\centering
\includegraphics[width=0.98\textwidth]{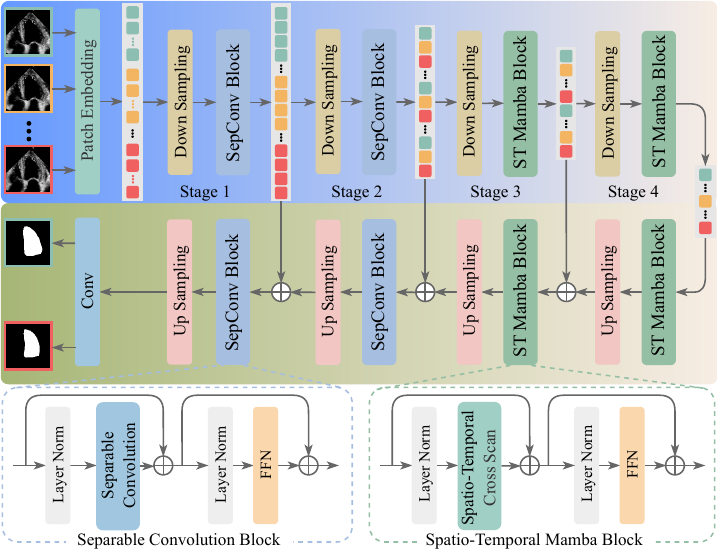}
\caption{Illustration of the proposed~\ourmodel~framework, which is a symmetric Encoder-Decoder architecture.}
\label{fig2}
\end{figure*}

\section{Methodology}

\textbf{Overview}. The architecture of our~\ourmodel, illustrated in Fig.~\ref{fig2}, consists of two modules: Encoder and Decoder. The first and second stages of the Encoder are primarily composed of stacked separable convolution blocks for low-level feature capture within single-frame images. The third and fourth stages are mainly composed of stacked spatio-temporal Mamba blocks for high-level feature capture across multiple frames.  
The Decoder architecture is symmetric to the Encoder, with a different number of stacked blocks. It is used to fuse multi-scale features and predict the segmentation mask.
Specifically, given a video clip of $T$ frames, denoted as $\mathbf{V}=\left \{  I_1, I_2, ..., I_T \right \}$, we first apply patch embedding to divide these frames into different patches. The patch sequence is then fed into the encoder, resulting in the $i$-th stage feature $\mathbf{F_i}$, with a size of $\frac{H}{2^{i+1}} \times \frac{W}{2^{i+1}}$, where $H$ and $W$ represent the height and width of the original frames, respectively, and $i\in \left \{ 1, 2, 3, 4 \right \}$. Finally, these multi-scale features are passed to the decoder, where operations such as inter-frame perception, intra-frame perception, and up-sampling are performed to generate the predicted segmentation results.

\begin{figure}[!t]
\centering
\includegraphics[width=0.98\columnwidth]{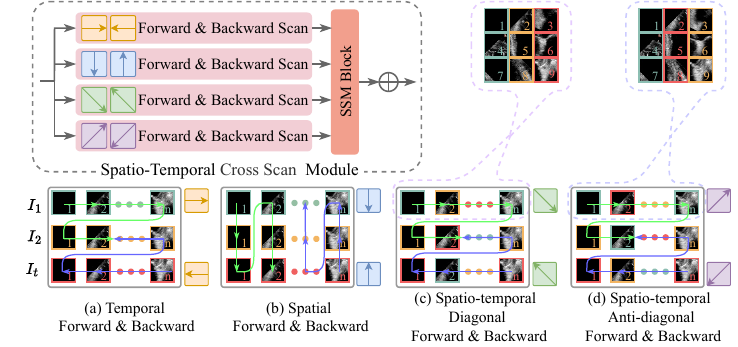} 
\caption{Spatio-temporal cross scan module and its scanning sequence ($I_t$ denotes the $t\text{-}th$ frame in the video clip, and $n$ denotes the number of patch sequences).}
\label{fig3}
\end{figure}

\subsection{Separable Convolution Block}
Fig.~\ref{fig2} illustrates the structure of the separable convolution block. For the $i$-stage feature embedding of low-level $\mathbf{F_i} \in \mathbb{R}^{T\times C_i\times H_i\times W_i}, i\in \left \{1, 2 \right \}$ of the given video clip, layer normalization $(\mathcal{LN(\cdot)})$ is performed before performing separable convolution operations $(\mathcal{SC}(\cdot))$. We follow the inverted separable convolution module from MobileNetV2~\cite{sandler2018mobilenetv2}. Subsequently, feeding the result into the FeedForward Network $(\mathcal{FFN(\cdot)})$ layer to capture low-level features such as edges and textures in single-frame images. The output features are either passed to the next stacked block or forwarded to the next stage. This process can be expressed as:
\begin{equation} \label{STIMB}
    \mathbf{F_i} = \mathcal{SC}(\mathcal{LN}(\mathbf{F_i})) + \mathbf{F_i}, \quad
    \mathbf{F_i} = \mathcal{FFN}(\mathcal{LN}(\mathbf{F_i})) + \mathbf{F_i}.
\end{equation}

\subsection{Spatio-temporal Mamba Block}
The structure of the spatial-temporal Mamba block is illustrated in Fig.~\ref{fig2}. For the $i$-stage feature embedding of high-level $\mathbf{F_i} \in \mathbb{R}^{T\times C_i\times H_i\times W_i}, i\in \left \{3, 4 \right \}$ of the given video clip, we transpose the channel and time dimensions and flatten the spatio-temporal feature embedding into a one-dimensional long sequence $\mathbf{s_i} \in \mathbb{R}^{C_i\times T H_i W_i}$. This sequence $\mathbf{s_i}$ is then fed into the spatio-temporal cross scan $(\mathcal{STCS(\cdot)})$ module and FFN layers. The STCS module establishes long-range dependencies both inter-frame and intra-frame from different spatio-temporal perspectives. This process can be defined as follows:  
\begin{equation} \label{STIMB}
    \mathbf{s_i} = \mathcal{STCS}(\mathcal{LN}(\mathbf{s_i})) + \mathbf{s_i}, \quad
    \mathbf{s_i} = \mathcal{FFN}(\mathcal{LN}(\mathbf{s_i})) + \mathbf{s_i}.
\end{equation}
Finally, the output feature sequences are reshaped back to their original shape, and after down-sampling, the feature embeddings are passed to the next stage.

\textbf{Spatio-temporal Cross Scan Module: }
As shown in Fig.~\ref{fig3}, the STCS module with state space model (S6)~\cite{gu2023mamba} is designed for the spatio-temporal sequence modeling of video frames. It selectively scans the input sequence from various spatio-temporal perspectives, capturing intricate spatio-temporal relationships and providing a comprehensive understanding of the context.

To better understand and explore the spatio-temporal relationships among frames, we first unfold each frame's patches into sequences along rows and columns. As illustrated in Fig.~\ref{fig3}, the patches of each frame are unfolded along rows to form temporal sequences, while the patches at the same position in different frames are unfolded along columns to form spatial sequences. The STCS module offers four different scanning modes: temporal, spatial, spatio-temporal diagonal, and spatio-temporal anti-diagonal. As depicted in Fig.~\ref{fig3}(a), the module scans simultaneously along the temporal sequence in both forward and backward directions to explore bidirectional temporal dependencies. In Fig.~\ref{fig3}(b), the module scans simultaneously along the spatial sequence in both upward and downward directions to explore bidirectional spatial dependencies. The selective spatio-temporal scanning explicitly considers both intra-frame and inter-frame coherencies and leverages the SSM to establish long-range dependencies of intra-frame and inter-frame. 

The heart motion during a heartbeat is not fully synchronous, leading to similarities and differences in information among video frames. To disrupt local data correlations, integrate global information, capture diverse features, and enhance the model's generalization and understanding of cardiac structure and motion, we propose a new spatio-temporal diagonal and anti-diagonal scanning method (see Fig.~\ref{fig3}(c) and (d)). We rearrange spatial sequence positions into diagonal and anti-diagonal patterns and scan temporally in forward and backward directions. This cross-frame and cross-position scanning improves the model's ability to integrate global information and understand cardiac motion and structure.

\textbf{Loss Function: }
During training, our loss function includes the Dice loss $\mathcal{L}_{dice}$~~\cite{milletari2016v} and binary cross-entropy loss $\mathcal{L}_{bce}$. Thus, the total loss function $\mathcal{L}_{total} = \alpha \mathcal{L}_{dice}(P, G) + (1-\alpha) \mathcal{L}_{bce}(P, G),$
where $G$ denotes the ground-truth, $P$ denotes the predicted masks, and the balance weight $\alpha = 0.8$ in our experiments.

\begin{table}[!t]
    \centering
    \footnotesize
    \setlength\tabcolsep{6.4pt}
    \caption{The quantitative results on the CAMUS. FLOPs represent the average computational complexity per frame at a size of $256 \times 256$.}
    \begin{tabular}{r|cc|cccc}
    \toprule
    Methods & Params & FLOPs & corr &bias$\pm$ std  & Dice & HD95\\ 
    \midrule
    UNet++~\cite{zhou2019unet++}       & 26.9M  & 37.7G & 81.68 & 6.05±6.81 & 91.87 & 16.16 \\
    TransUNet~\cite{chen2021transunet} & 105.3M & 38.6G & 86.22 & \textBF{1.72±6.07} & 92.73 & 13.71 \\
    SegFormer~\cite{xie2021segformer}  & 47.4M  & 20.9G & 79.10 & 7.84±7.21 & 91.17 & 18.46 \\
    H2Former~\cite{he2023h2former}     & 33.7M  & 33.1G & 82.35 & 5.79±6.87 & 91.78 & 16.06 \\
    \midrule
    SSCF~\cite{SSCF}                   & 53.7M  & 15.1G & 84.48 & 4.11±6.37 & 92.59 & 14.18 \\
    PKEchoNet~\cite{wu2023super}       & 25.7M  & 7.2G  & 76.20 & 4.13±8.45 & 93.02 & 12.93 \\
    VideoMamba~\cite{li2024videomamba} & 75.6M  & 22.0G & 75.21 & 8.02±8.00 & 91.53 & 16.43 \\
    Vivim~\cite{yang2024vivim}         & 59.6M  & 20.6G & 78.09 & 5.75±7.35 & 92.79 & 12.74 \\
    \ourmodel~(Ours)                   & 31.2M  & 5.6G & \textBF{90.47} & 2.43±5.02 & \textBF{93.89} & \textBF{11.29} \\ 
    \bottomrule
    \end{tabular}
    \label{table1}
\end{table}

\begin{table*}[!t]
    \centering
    \footnotesize
    \setlength\tabcolsep{2.0pt}
    \caption{The quantitative results on the EchoNet-Pediatric and EchoNet-Dynamic datasets. The HD95 metrics are reported in pixels.}
    \begin{tabular}{r|cccc|cccc}
    \toprule
    \multirow{2}{*}{Methods}    & \multicolumn{4}{c|}{EchoNet-Pediatric} & \multicolumn{4}{c}{EchoNet-Dynamic} \\ 
                                & \multicolumn{1}{c}{corr} & \multicolumn{1}{c}{bias$\pm$ std}  & \multicolumn{1}{c}{Dice} & \multicolumn{1}{c|}{HD95}  
                                & \multicolumn{1}{c}{corr} & \multicolumn{1}{c}{bias$\pm$ std}  & \multicolumn{1}{c}{Dice} & \multicolumn{1}{c}{HD95}  \\ 
    \midrule
    UNet++~\cite{zhou2019unet++}       & 69.33 & 7.62±10.16 & 90.73 & 3.65 & 73.91 & 9.43±9.02 & 91.50 & 2.99 \\
    TransUNet~\cite{chen2021transunet} & 73.09 & 6.54±9.81  & 91.11 & 3.52 & 74.17 & 4.67±9.51 & 91.92 & 2.96 \\ 
    SegFormer~\cite{xie2021segformer}  & 66.72 & 6.25±10.77 & 91.10 & 3.52 & 73.12 & 7.07±9.37 & 92.07 & 2.90 \\
    H2Former~\cite{he2023h2former}     & 69.77 & 6.24±10.06 & 90.89 & 3.58 & 74.78 & 6.10±9.23 & 91.68 & 3.12 \\ 
    \midrule
    SSCF~\cite{SSCF}                   & 63.29 & 5.28±11.82 & 91.07 & 3.50 & 74.87 & 6.26±9.16 & 92.35 & 2.80 \\
    PKEchoNet~\cite{wu2023super}       & 65.04 & 6.23±11.21 & 91.00 & 3.57 & 75.43 & 4.34±9.50 & 92.45 & 2.71 \\
    VideoMamba~\cite{li2024videomamba} & 67.34 & 6.39±11.39 & 91.06 & 3.50 & 78.62 & 4.50±8.29 & 92.48 & 2.75 \\
    Vivim~\cite{yang2024vivim}         & 69.92 & 5.59±10.31 & 91.12 & 3.46 & 81.12 & 7.02±7.47 & 92.46 & 2.73 \\
    \ourmodel~(Ours)                   & \textBF{76.91} & \textBF{1.29±8.68}  & \textBF{91.90} & \textBF{3.23} & \textBF{84.50} & \textBF{0.95±6.75} & \textBF{92.67} & \textBF{2.66}   \\ 
    \bottomrule
    \end{tabular}
    \label{table2}
\end{table*}

\section{Experiments}
\textbf{Datasets: }
In this study, three publicly available echocardiography video datasets are used, namely CAMUS~\cite{leclerc_deep_2019}, EchoNet-Pediatric~\cite{reddy_videobased_2023}, and EchoNet-Dynamic~\cite{yang_graphecho_2023}. 
$\bullet$~\textbf{CAMUS} comprises 500 cases acquired at the University Hospital of St Etienne (France), each including 2D apical 2-chamber and 4-chamber view videos, with annotations provided for all frames.
$\bullet$~\textbf{EchoNet-Pediatric} is collected from Lucile Packard Children’s Hospital Stanford, including 7,643 video clips from 1,958 patients aged 0 to 18 years. This dataset consists of either parasternal short axis or apical 4-chamber views, with only the ED and ES frames annotated. 
$\bullet$~\textbf{EchoNet-Dynamic} consists of 10,030 apical 4-chamber view echocardiography videos collected from Stanford University Hospital, with only ED and ES frames annotated for each video. 
We uniformly sample 10 frames of each video clip from datasets, following previous research~\cite{Deng_2024_CVPR,wu2023super}. The video clips are cropped to ensure that the ED frame is the first and the ES frame is the last, thereby capturing a complete heartbeat cycle. The frame size was adjusted to $256 \times 256$, and only the annotations from the ED and ES frames were used for training and evaluation. 
For the EchoNet-Dynamic dataset, we used the original dataset splits. For the other datasets, we follow related studies~\cite{yang_graphecho_2023,ye2024p} and split the data into training, validation, and testing sets in an 8:1:1 ratio.

\textbf{Evaluation Metrics: }
We report three statistical metrics for left ventricular ejection fraction. 
The estimation methods vary due to different views provided by the datasets. 
For the EchoNet-Dynamic and EchoNet-Pediatric datasets, both ground truth and predicted ejection fractions are obtained using the Simpson's single-plane method of disks~\cite{leclerc_deep_2019}.
For the CAMUS dataset, the Simpson's biplane method of disks~\cite{yang_graphecho_2023} is used to calculate ejection fractions.
We follow~\cite{Deng_2024_CVPR,wu2023super} and calculate the Pearson correlation coefficient (corr), mean bias (bias), and standard deviation (std) for the predicted and ground truth ejection fractions.
Additionally, we employed two widely used segmentation evaluation metrics: the mean Dice coefficient (Dice) and Hausdorff Distance at 95\% (HD95).

\textbf{Implementation Details:}
Our model is implemented using PyTorch and is trained or inferred on two NVIDIA RTX 4090 GPU. We train our model end-to-end using the Adam optimizer and employ the cosine annealing strategy to adjust the learning rate. The maximum and minimum learning rates are set to 1e-4 and 1e-5, respectively, and the maximum training epoch is set to 120. 
During training, we apply gamma augmentation, random scaling, random rotation, and random contrast adjustments, each with a probability of 0.5.

\begin{figure}[!t]
\centering
\includegraphics[width=1.0\textwidth]{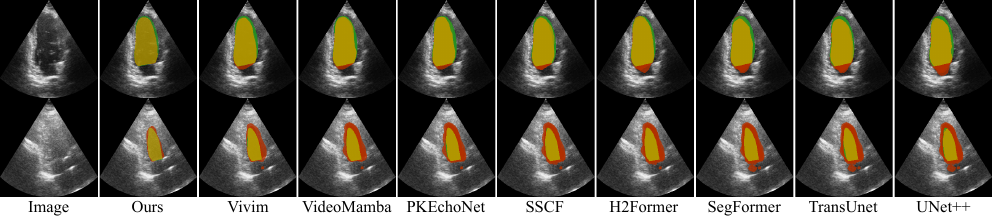} 
\caption{Visualization of segmentation maps for different models. }
\label{fig4}
\end{figure}

\subsection{Comparison with State-of-the-art Methods}

\textbf{Quantitative Comparisons: }
The quantitative results for the three datasets are presented in Tables~\ref{table1} and~\ref{table2}. It can be seen that our model outperforms other methods in all datasets. However, image-based methods (The first four) still exhibit competitive performance on certain datasets. These methods focus on capturing local features within single-frame images, whereas video-based methods (The last four), although proficient in capturing temporal information, may not fully exploit their advantages on datasets with minimal inter-frame differences or subtle dynamic changes. 
Our model hierarchically processes single-frame and multi-frame information, balancing both detailed features and dynamic changes in echocardiography. This approach is particularly advantageous for ejection fraction estimation. Additionally, it maintains an optimal balance between model performance and computational complexity.

\textbf{Qualitative Comparisons:}
We present visualizations of several challenging cases. As shown in Fig.~\ref{fig4}, these sample images exhibit artifacts, speckle noise, and blurred boundaries. Such challenging conditions mislead most of the compared models, resulting in missed or misclassified regions. In contrast, our model accurately locates the regions and delineates the boundaries. These visualizations further demonstrate that our approach can achieve better segmentation results and robustly handle poor-quality images.

\subsection{Ablation Study}

\begin{table*}[!t]
    \centering
    \footnotesize
    \setlength\tabcolsep{1.2pt}
    \caption{Quantitative results of ablation studies.}
    \begin{tabular}{l|cccc|cccc}
    \toprule
    \multirow{2}{*}{Settings}    & \multicolumn{4}{c|}{CAMUS} & \multicolumn{4}{c}{EchoNet-Dynamic} \\
                                & \multicolumn{1}{c}{corr} & \multicolumn{1}{c}{bias$\pm$ std}  & \multicolumn{1}{c}{Dice} & \multicolumn{1}{c|}{HD95}  
                                & \multicolumn{1}{c}{corr} & \multicolumn{1}{c}{bias$\pm$ std}  & \multicolumn{1}{c}{Dice} & \multicolumn{1}{c}{HD95}  \\ 
    \midrule
    Image-level          & 83.48 & 5.28±6.71 & 92.93 & 13.91 & 74.79 & 6.00±9.34 & 91.93 & 2.97 \\
    Video-level          & 80.67 & 4.48±7.04 & 93.04 & 13.09 & 78.01 & 4.62±7.92 & 92.03 & 2.91 \\ 
    \midrule
    w/o Temporal         & 83.73 & 4.20±6.67 & 93.02 & 14.26 & 78.42 & 4.80±8.02 & 92.10 & 2.90 \\
    w/o Spatio           & 80.44 & 5.05±7.44 & 93.26 & 12.57 & 77.86 & 4.76±8.21 & 92.32 & 2.79 \\
    w/o ST Diagonal      & 86.69 & 4.05±5.99 & 93.21 & 12.40 & 79.65 & 6.03±7.70 & 92.12 & 2.84 \\
    w/o ST Anti-diagonal & 88.09 & 2.95±5.57 & 93.27 & 12.28 & 81.11 & 4.85±7.54 & 92.26 & 2.82 \\ 
    \midrule
    \ourmodel~(Ours)     & \textBF{90.47} & \textBF{2.43±5.02} & \textBF{93.89} & \textBF{11.29} & \textBF{84.50} & \textBF{0.95±6.75} & \textBF{92.67} & \textBF{2.66} \\ 
    \bottomrule
    \end{tabular}
    \label{table3}
\end{table*}

\textbf{Effectiveness of Hierarchical Design: }
To validate the effectiveness of the hierarchical design, we perform two sets of ablation experiments. One set processes single-frame images at all stages using only separable convolution blocks (labeled as Image-level). The other set processes multi-frame images at all stages using only spatio-temporal Mamba blocks (labeled as Video-level).  
As shown in Table~\ref{table3}, the model performance in both experiments decreased to varying degrees, especially in the key metric of Pearson correlation for ejection fraction estimation. This demonstrates that by extracting fine-grained details at the low level and modeling cross-frame temporal relationships at the high level, the model effectively handles subtle differences across different conditions, providing a more reliable foundation for clinical assessment of cardiac function.

\textbf{Effectiveness of Spatio-temporal Cross Scan Module: }
We investigate the effect of each mode in the STCS module. As shown in Table~\ref{table3}, all performance metrics exhibit varying degrees of decline compared to our full method. Temporal scanning captures sequential dynamic information during cardiac motion, which is crucial to accurately modeling heart movement patterns. Spatial scanning aids the model in understanding changes at the same location across different time frames. This enhances the model's ability to perceive spatial consistency features. Spatio-temporal diagonal and anti-diagonal scanning effectively capture complex interactions between temporal and spatial dimensions, enhancing the model's ability to integrate spatio-temporal information. 

\section{Conclusion}
We propose a novel method~\ourmodel~that employs a hierarchical design. The low-level stages use convolutions to extract local details from single-frame images, while the high-level stages leverage the Mamba architecture to process spatio-temporal information across multiple frames. By handling single-frame and multi-frame information hierarchically, the model's accuracy and robustness are enhanced. Extensive experimental results demonstrate that our method achieves state-of-the-art results on three benchmark datasets. 

% \subsubsection{Acknowledgement.} This work was in part supported by the National Science Fund of China under Grant Nos. 62172228, U24A20330, and 62361166670.

% \subsubsection{Disclosure of Interests.} The authors have no competing interests to declare that are relevant to the content of this article.

\bibliographystyle{splncs04}
\bibliography{Paper-2745}

\end{document}